\documentclass{article}

\usepackage{arxiv}

\usepackage[utf8]{inputenc} 
\usepackage[T1]{fontenc}    
\usepackage{hyperref}       
\usepackage{url}            
\usepackage{booktabs}       
\usepackage{amsfonts}       
\usepackage{nicefrac}       
\usepackage{microtype}      
\usepackage{lipsum}
\usepackage{graphicx}
\usepackage{float}
\graphicspath{ {./images/} }

\title{Ruppert–Polyak averaging for Stochastic Order Oracle}

\author{
 Smirnov V.N.$^{* 1}$ \\
 smirnov.vladimir@phystech.edu\\
   \And
 Kazistova K.M.$^{* 1}$ \\
 kazistova.km@phystech.edu\\
  \And
 Sudakov I.A.$^{* 2}$ \\
 iasudakov@edu.hse.ru\\
  \And 
  \\
  \And
 Leplat V.$^{3}$ \\
 v.leplat@innopolis.ru\\
  \And
 Gasnikov A.V.$^{1, 3, 4, 5}$ \\
 gasnikov@yandex.ru\\
  \And
 Lobanov A.V.$^{1, 6, 7}$ \\
 lobbsasha@mail.ru\\
}

\begin{document}

\maketitle
\def\thefootnote{*}\footnotetext{These authors contributed equally}
\def\thefootnote{1}\footnotetext{Moscow Institute of Physics and Technology}
\def\thefootnote{2}\footnotetext{Higher School of Economics}
\def\thefootnote{3}\footnotetext{Innopolis University}
\def\thefootnote{4}\footnotetext{Caucasus Mathematical Center, Adyghe State University}
\def\thefootnote{5}\footnotetext{Steklov Mathematical Institute of Russian Academy of Sciences}
\def\thefootnote{6}\footnotetext{Skolkovo Institute of Science and Technology}
\def\thefootnote{7}\footnotetext{ISP RAS Research Center for Trusted Artificial Intelligence}

\begin{abstract}
Black-box optimization, a rapidly growing field, faces challenges due to limited knowledge of the objective function's internal mechanisms. One promising approach to address this is the Stochastic Order Oracle Concept. This concept, similar to other Order Oracle Concepts, relies solely on relative comparisons of function values without requiring access to the exact values. 
This paper presents a novel, improved estimation of the covariance matrix for the asymptotic convergence of the Stochastic Order Oracle Concept. Our work surpasses existing research in this domain by offering a more accurate estimation of asymptotic convergence rate. Finally, numerical experiments validate our theoretical findings, providing strong empirical support for our proposed approach.
\end{abstract}


\section{Introduction}
\setcounter{equation}{0}

Optimization is the cornerstone of machine learning, powering the development of complex algorithms like deep neural networks. While traditional methods like stochastic gradient descent (SGD) and ADAM have achieved remarkable success, their reliance on gradient calculations poses limitations in certain scenarios. This arises in areas like black-box adversarial attacks on neural networks~\cite{chen2017zoo, papernot2017practical} and policy search within reinforcement learning. These approaches rely solely on function evaluations, circumventing the need for gradient calculations~\cite{conn2009introduction, larson2019derivative}.

In scenarios where humans provide feedback for reinforcement learning, zeroth-order oracles prove particularly useful. Humans often struggle to quantify their feedback numerically, but readily discern which option is superior. Recent research, e.g.~\cite{tang2023zeroth, ouyang2022training} demonstrates that leveraging even the signs of gradient descent can yield impressive results, exemplified by the remarkable success of large language models (LLMs) trained using reinforcement learning from human feedback (RLHF). Furthermore, in various optimization applications, it may be prohibitively resource-intensive to compute exact values of a target function with high precision. In such cases, the ability to compare two values without requiring explicit calculations can lead to significant resource savings, enhancing the efficiency of the optimization process.

These advances in zeroth-order optimization and preference-based learning hold promise for pushing the boundaries of machine learning, enabling us to build even more sophisticated and powerful algorithms.

In this paper, we consider a convex optimization problem in the following form:
$$
\min\limits_{x\in\mathbb{R}^d}\left\{ f(x)=\mathbb{E}_{\xi \sim \mathcal{D}} f_\xi(x)\right\}
$$
where $f:\mathbb{R}^d\rightarrow\mathbb{R}$ is a $\mu$-strongly convex function.

This problem setup has wide applicability in machine learning, particularly in tasks like empirical risk minimization. In this context, $\mathcal{D}$ represents the distribution of training data points, and $f_\xi(x)$ corresponds to the loss incurred by the model when applied to a specific data point, denoted by $\xi$.

Our work is dedicated to Stochastic Order Oracle Concept in optimization algorithms. Such an oracle arises when the outcome of comparing function values is influenced by some random noise. More specifically, we focus on the examination of an algorithm that utilizes a stochastic order oracle, as presented in the work of Lobanov et al.~\cite{lobanov2024order}. For this algorithm, an estimate of its asymptotic convergence rate has been derived. In our study, we provide a new asymptotic estimate for the convergence rate of the aforementioned algorithm. These estimates are compared both theoretically and empirically, and our estimate proves to be superior.

\subsection{Our Contribution}
Our contributions are the following:
\begin{itemize}
    \item Building on the recently introduced concept of \textit{stochastic order oracle} (see Section \ref{sec3} for more details), we extend the work of Lobanov et al.~\cite{lobanov2024order}. In particular, by incorporating the Ruppert–Polyak averaging approach into our algorithm, we demonstrate that, under mild conditions, the quantity \(\sqrt{k}(\overline{x}_k - x^*)\) is asymptotically normal. That is, we show that \(\sqrt{k}(\overline{x}_k - x^*) \sim \mathcal{N}(0,V)\), where \(V\) is the covariance matrix that proves to be more accurate than the one proposed in \cite{lobanov2024order}.
    \item We derive an optimal expression for the initial step size \(\eta\) that minimizes any unitarily invariant norm of the covariance matrix \(V\), thereby achieving minimal dispersion of the distribution of limit points around the minimizer \(x^\star\) of the objective function \(f\).
    \item We conduct comprehensive numerical experiments which support our theoretical findings. 
\end{itemize}
\subsection{Paper Organization}
The paper has the following organization: in section \ref{sec3} we describe Stochastic Order Oracle Concept and an algorithm that employs this concept, then we derive a new estimation for the asymptotic convergence rate of this algorithm and compare it with the existing one. Section \ref{sec4} is devoted to numerical experiments that confirm theoretic results. 

\subsection{Main assumptions and notations}
Before discussing related works, we present the notation and main assumptions we use in our work.

\textbf{Notation.} We use $\langle x, y \rangle := \sum\limits_{i=1}^{d} x_i y_i$ to denote standard inner product of $x, y \in \mathbb{R}^d$. We denote Euclidean norm in $\mathbb{R}^d$ as $\|x\| := \sqrt{\sum\limits_{i=1}^{d} x_i^2.}$ We use $\mathbf{e}_i \in \mathbb{R}^d$ to denote the $i$-th unit vector. We denote by $\nabla f(x)$ the full gradient and by $\nabla^2 f(x)$ the Hessian of function $f$ at point $x \in \mathbb{R}^d$. We use $S^d(r) := \{x \in \mathbb{R}^d: \|x|\ = r\}$ to denote Euclidean sphere. We denote by $x^*$ the minimum point of function $f$. For matrices $A, B$ $A \succeq  B$ means $(A - B)$ is positive semi-definite matrix. 

For all our theoretical results, we assume that $f(x)$ is $L$-smooth and $\mu$-strongly convex:

\begin{assum}
A function $f: \mathbb{R}^d \rightarrow \mathbb{R}$ is $L$-smooth w.r.t. the norm $\|\cdot\|$, if for any $x, y \in \mathbb{R}^d$ the following inequality holds:
$$\|\nabla f(x) - \nabla f(y)\| \leq L\|x-y\|.$$
\end{assum}

\begin{assum}
    A function $f: \mathbb{R}^d \rightarrow \mathbb{R}$ is $\mu \geq 0$ strongly convex w.r.t. the norm $\|\cdot\|$, if for any $x, y \in \mathbb{R}^d$ the following inequality holds:
    $$f(y) \geq f(x) + \langle \nabla f(x), y - x \rangle + \frac{\mu}{2} \|y - x\|^2.$$
\end{assum}

\subsection{Related Works}\label{sec2}

Our work is grounded in the results of the work of B.T. Polyak and A.B. Juditsky~\cite{polyak1992acceleration}, in which the asymptotic convergence rate of optimization algorithms plays significant role. As stated in their study, it is often possible to demonstrate that $\sqrt{k}(x_k - x^*)$ is asymptotically normal with mean equal to zero and some covariance matrix. (Here $k$ is the number of optimization algorithm step, $x^*$ is true point of minimum.) This covariance matrix serves as an adequate descriptor of the behavior of the recurrent process for large $k$. This characterization is comprehensive, as it is well-known that a normal distribution is entirely described by its mean and covariance matrix. If it is found that the covariance matrix for one algorithm is smaller (in the matrix sense) than for another, then any "reasonable" scalar measure of accuracy for the first algorithm would also be superior. 

In our work, we consider an algorithm employing a stochastic order oracle, as introduced in the work of Lobanov et al.~\cite{lobanov2024order}. This study has derived the asymptotic covariance matrix for $\sqrt{k}(x_k - x^*)$. This result builds upon the works of Saha et al.~\cite{saha2021dueling} and Polyak, Tsypkin~\cite{polyak1980optimal}. We obtain a formula for the asymptotic covariance matrix of $\sqrt{k}(\overline{x}_k - x^*)$, where $\overline{x}_k = \sum\limits_{i=0}^{k-1} x_i$, thereby providing a new estimate for the asymptotic convergence rate. Our findings are based on the research~\cite{polyak1992acceleration}. 

We compare the covariance matrix formula we have derived with the one from the work~\cite{lobanov2024order}, and present both theoretical and experimental evidence demonstrating that our estimate is superior.

\section{Algorithm and Convergence Results}\label{sec3}
\subsection{Stochastic Order Oracle}

We are considering a concept of so-called Order Oracle (Lobanov et al.~\cite{lobanov2024order}): an oracle that can only compare two values of a target function. This concept is weaker than Zeroth-Order Oracle, because the values of the function are unknown. More formally, Order Oracle can return only the following value at two points $x$ and $y$:

$$
\psi(x, y) = \mathsf{sign}(f(x) - f(y))
$$

Actually, we can't talk about a precise Order Oracle without paying attention to some noise that can be applied to the target function somehow. For example, turning again to a RLHF application of such an oracle, people's feedback may slightly (or vastly) depend on their mood, biased opinion etc. We are taking it into account saying that we are working only with some realizations of a noisy target function $f(x,\xi)$: $f(x)=\mathbb{E}_\xi f(x,\xi)$. Under this condition the considering Order Oracle concept turns into a Stochastic Order Oracle concept, which can return only the following value:

\begin{equation}\label{eq1}
\phi(x, y, \xi) = \mathsf{sign}[f(x, \xi) - f(y, \xi)].
\end{equation}

Using this concept we can introduce the following optimization algorithm:

\begin{equation}\label{algorithm}
x_{k+1} = x_k - \eta_k \phi(x_k + \gamma \mathbf{e}_k, x_k - \gamma \mathbf{e}_k, \xi_k)\mathbf{e}_k,
\end{equation}

where $\gamma > 0$ is a smoothing parameter, $\mathbf{e}_k \in S^d(1)$ is a vector uniformly distributed on the Euclidean sphere, as in idea proposed in~\cite{nesterov2017random}, $\eta_k = \eta/k^\theta$, $\theta$ is chosen so that $\sum\limits_{k=1}^{+\infty}\eta_k=+\infty$ and $\sum\limits_{k=1}^{+\infty}\eta_k^2<+\infty$.

\subsection{Main Results}
\subsubsection{Base Algorithm's Convergence}

To establish the convergence results for the proposed algorithms in the case of (stochastic) convex optimization~\cite{boyd2004convex, nesterov2018lectures}, we first recall two useful lemmas taken from the work of Saha et al.~\cite{saha2021dueling} and used in the work of Lobanov et al.~\cite{lobanov2024order} :

\begin{lem}{\cite[Lemma 5.1]{lobanov2024order}}\label{l51}
     Let function $f$ be $L$-smooth, $\gamma_k = \frac{\|\nabla f(x_k, \xi_k)\|}{\sqrt{d}L}$, $\mathbf{e}_k\in S^d(1)$, then the following holds:
    $$
    \phi(x_k+\gamma_k \mathbf{e}_k,x_k-\gamma_k \mathbf{e}_k, \xi_k)\mathbf{e}_k =\mathsf{sign}\left[ \left\langle \nabla f(x_k,\xi_k), \mathbf{e}_k\right\rangle \right]\mathbf{e}_k.
    $$
\end{lem}

\begin{lem}{\cite[Lemma 2]{saha2021dueling}}\label{l52}
    Let vector $\nabla f(x,\xi)\in\mathbb{R}^d$ and vector $\mathbf{e}\in S^d(1)$, then with some constant $c$ we have
    $$
    \mathbb{E}_\mathbf{e}\left[ \mathsf{sign}\left[ \left\langle \nabla f(x,\xi), \mathbf{e} \right\rangle \right]\mathbf{e} \right] = \frac{c}{\sqrt{d}}\cdot \frac{\nabla f(x,\xi)}{\|\nabla f(x,\xi)\|}.
    $$
\end{lem}

From the work of Saha et al.~\cite{saha2021dueling} we know that $c \in [\frac{1}{20}, 1].$

This shows that proposed algorithm is a variation of normalized gradient descent~\cite{murray2019revisiting},~\cite{polyak1980optimal},~\cite{saha2021dueling}.

In further discussion, we will consider more specific and simpler type of noise -- additive to a gradient of the target function: $\nabla f(x, \xi) = \nabla f(x) + \xi$. Also we assume that the noise is bounded, i.e. $\|\xi\|<\Delta$. Too huge value of $\|\Delta\|$ breaks algorithm's \ref{algorithm} convergence to small values of $\varepsilon$, where $\|x_k - x^*\| < \varepsilon$. Influence of the magnitude of $\Delta$ is a separate research topic that is not covered in this paper. We assume that $\|\Delta\|$ is small enough to make it possible for the algorithm \ref{algorithm} to reach the desired accuracy.

This assumptions allow us to use conclusions made in the base work~\cite{lobanov2024order}.

Given the fact (from Lemmas \ref{l51} and \ref{l52}) that the direction in which the step of algorithm \ref{algorithm} is taken is the direction of normalized stochastic gradient descent and based on the work of Polyak and Tsypkin~\cite{polyak1980optimal} the following theorem holds:

\begin{teo}{\cite[Theorem 5.3]{lobanov2024order}}\label{t53}
    Let the function $f$ be $L$-smooth and $\mathbf{e}\in S^d(1)$, then for the algorithm \ref{algorithm} with step size $\eta_k=\eta/k$ the value $\sqrt{k}\left( x_k-x^* \right)$ is asymptotically normal: $\sqrt{k}\left( x_k-x^* \right)\sim\mathcal{N}\left( 0, V \right)$, where the matrix $V$ is as follows:
    \begin{equation}\label{eq_gasnikov}
      V=\frac{\eta^2}{d}\left( 2\eta(1-\frac{1}{d})\frac{c}{\sqrt{d}}\alpha\nabla^2f\left( 
x^* \right) - I \right)^{-1},  
    \end{equation}
    
    $\alpha=\mathbb{E}_\xi\left[ \|\xi\|^{-1} \right]<\infty, 2\eta(1-1/d)\frac{c}{\sqrt{d}}\alpha\nabla^2 f(x^*)>I$ (where $I$ is a unit matrix).
\end{teo}

\subsubsection{Parameter Optimization}

From Theorem \ref{t53} one can observe that $V$ may be expressed as a function $V(\eta):\mathbb{R}\rightarrow\mathbb{R}^{d\times d}$. 
To obtain the most concentrated distribution around \( x^* \), we need to determine the optimal parameter \( \eta \). 
This can be achieved by solving

$$\eta_0={\mathsf{argmin}_{\eta\in\mathbb{R}}}\{\|V(\eta)\|\}$$
for some matrix norm $\|\cdot\|$. The expression for the optimal \(\eta_0\) is presented in the Lemma~\ref{lem:opt_eta}.

\begin{lem}\label{lem:opt_eta}
    For an unitarily invariant norm $\|\cdot\|$ and $\mu$-strongly convex $f(x)$ $\exists!\ \eta_0={\mathsf{argmin}_{\eta\in\mathbb{R}}}\{\|V(\eta)\|\}=\frac{d\sqrt{d}}{(d-1)\cdot\alpha c\cdot\mu}$.
\end{lem}

Let's denote $V_0=V(\eta_0)$. Our goal is to improve the estimation of this covariance matrix, i.e. find some random vector as a function of sample $X_k$ that will be distributed as $\mathcal{N}(0, V^{\prime})$: $V_0\succeq V^{\prime}$ with the meaning of $V_0-V^{\prime}$ is a positive semi-definite matrix.

\subsubsection{Application of results of B.T. Polyak and A.B. Juditsky}

We begin this section by presenting several key assumptions, after which we present our main convergence results in Theorem~\ref{t1}.

\begin{assum}\label{as1}
Let function $f(x)$ be $\mu$-strongly convex ($\mu I \leq \nabla^2f(x)$), $L$-smooth\\ ($\nabla^2f(x)~\leq~LI$) and twice continuously differentiable for all $x$ and some $\mu > 0$ and $L > 0$,\\ $\nabla^2 f(x^*) > 0$; here $I$ is the identity matrix.
\end{assum}

\begin{assum}\label{as2}
$(\xi_k)_{k \geq 0}$ is the sequence of mutually independent and identically distributed random variables. $\xi_k$ distribution has spherical symmetry.
\end{assum}

To improve convergence properties of the algorithm \ref{algorithm}, we use the Ruppert-Polyak averaging procedure~\cite{gadat2023optimal}, that consists in introducing average over the past iterations of algorithm. We denote such averaging as: 
\begin{equation}
\overline{x}_k = \frac{1}{k} \sum_{i=0}^{k-1}x_i
\end{equation}

\begin{teo}\label{t1}
Let Assumptions \ref{as1}-\ref{as2} be fulfilled. Then for algorithm \ref{algorithm} the value $\sqrt{k}(\overline{x}_k - x^*)$  is asymptotically normal: $\sqrt{k}(\overline{x}_k - x^*) \sim \mathcal{N}(0, V)$, where the matrix $V$ is as follows:

\begin{equation}\label{eq_ours}
V = \frac{d}{(d-1)^2\alpha^2}\nabla^2f(x^*)^{-2},
\end{equation}

$\alpha =\int {\|z\|}^{-1}dP(z) < \infty$, where $P(z)$ is distribution function of stochastic noise $\xi$.
\end{teo}

This result has several advantages over the one from Theorem \ref{t53}:

\begin{itemize}
    \item Covariance matrix depends on the Hessian in the $-2$nd degree, not $-1$st, what narrows the distribution of the corresponding random vector.
    \item The result doesn't depend on the unknown constant $c$.
\end{itemize}

Thereby, applying Ruppert-Polyak averaging procedure~\cite{gadat2023optimal} to the algorithm \ref{algorithm} will demonstrate better asymptotic convergence that also may be estimated more precisely due to the reduction of the number of parameters.

To prove the Theorem \ref{t1}, firstly denote a function $\varphi(x): \mathbb{R}^d \rightarrow \mathbb{R}^d$,

\begin{equation}
\varphi(x) = \frac{c}{\sqrt{d}} \cdot \frac{x}{\|x\|}.
\end{equation}

Secondly, let's formulate Lemma \ref{l1}, that will continue the proof.

\begin{lem}\label{l1}
Under the conditions of Theorem \ref{t1} the following statements are fulfilled. 

\begin{stat}\label{st1}
It holds that $\|\phi(x, y, \xi) \mathbf{e}\| \leq K_1(1 + \|\xi\|)$.
\end{stat}

\begin{stat}\label{st2}

The function $\psi(x) = \mathbb{E}_{\xi} \varphi (x + \xi)$ is defined and has a derivative at zero, $\psi(0) = 0$ and $x^T\psi(x) > 0$ for all $x \neq 0$. Moreover, there exist $\varepsilon, K_2 > 0, 0 < \lambda \leq 1$, such that $$\|\psi^{\prime}(0)x - \psi(x)\| \leq K_2\|x\|^{1 + \lambda}$$

for $\|x\| < \varepsilon$.
\end{stat}

\begin{stat}\label{st3}

The matrix function $\chi(x) = \mathbb{E}_{\xi} \varphi (x + \xi) \varphi (x + \xi)^T$ is defined and is continuous at zero.
\end{stat}

\begin{stat}\label{st4}

The matrix $-G = -\psi^{\prime}(0) \nabla^2f(x^*)$ is Hurwitz, i.e., Re $\lambda_i (G) > 0, i = \overline{1, d}$.
\end{stat}
\end{lem}

Lemma \ref{l1} allows us to apply asymptotic convergence results from the work of B.T. Polyak and A.B. Juditsky~\cite{polyak1992acceleration} to the algorithm \ref{algorithm} what completes the proof of the Theorem \ref{t1} from this work.

For a detailed proof of Theorem \ref{t1} and Lemma \ref{l1} see Section \ref{proofs}.

\subsubsection{Convergence comparison}

\begin{lem}\label{lem:better-estimation}
    Under conditions of Theorems \ref{t53}, \ref{t1} and Lemma \ref{lem:opt_eta}, for the
    $$
    \eta_0={\mathsf{argmin}_{\eta\in\mathbb{R}}}\left\{\left\|\frac{\eta^2}{d}\left( 2\eta(1-\frac{1}{d})\frac{c}{\sqrt{d}}\alpha\nabla^2f\left(x^* \right) - I \right)^{-1}\right\|\right\}
    $$
    The following inequality holds:
    $$
    V(\eta_0)=\frac{\eta_0^2}{d}\left( 2\eta_0(1-\frac{1}{d})\frac{c}{\sqrt{d}}\alpha\nabla^2f\left(x^* \right) - I \right)^{-1}\succeq \frac{d}{(d-1)^2\alpha^2}\nabla^2f(x^*)^{-2}
    $$
\end{lem}

Lemma \ref{lem:better-estimation} illustrates the better convergence rate of averaging procedure. The proof is given in appendix.

\section{Numerical Experiments}\label{sec4}

In this section, we perform several experiments to validate the theoretical results presented in the previous sections. For this purpose, we have established the following test setup: we executed algorithm~\eqref{algorithm} \(10,000\) times with different initial points across three settings:

\begin{itemize}
    \item \textbf{Setting 1:} We set \(\eta = 5\) and report the empirical distribution of \(\sqrt{k}(x_k - x^\star)\).
    \item \textbf{Setting 2:} We set \(\eta = \eta_0\) (the optimal value derived in Lemma \ref{lem:opt_eta}) and report the empirical distribution of \(\sqrt{k}(x_k - x^\star)\).
    \item \textbf{Setting 3:} We report the empirical distribution of \(\sqrt{k}(\bar{x}_k - x^\star)\), including to algorithm~\eqref{algorithm} the computation of the averaged sequence $\bar{x}_k=\frac{1}{k} \sum_{i=0}^{k-1}x_i$.
\end{itemize}

Note that from the first setting we calculated an estimation of the product $c\cdot\alpha$, which is used to compute an optimal $\eta_0$ (see appendix for details).

Further, the experiments were conducted on a standard quadratic optimization problem represented as $f(x) = \frac{1}{2}x^TAx + bx + c$, where $A \in S_{++}^n$, that is $A$ is a positive definite matrix of size $n \times n$. In this context, the constants $\mu$ and $\L$ correspond to the minimum and maximum eigenvalues of $A$, respectively. To facilitate the illustration of histograms related to the distribution of values derived from the theorems, we considered the case $n=2$. 

\begin{figure}[H]
\begin{minipage}{.33\textwidth}
    \vspace{-6pt}
    \centering
    \includegraphics[width=0.98\textwidth]{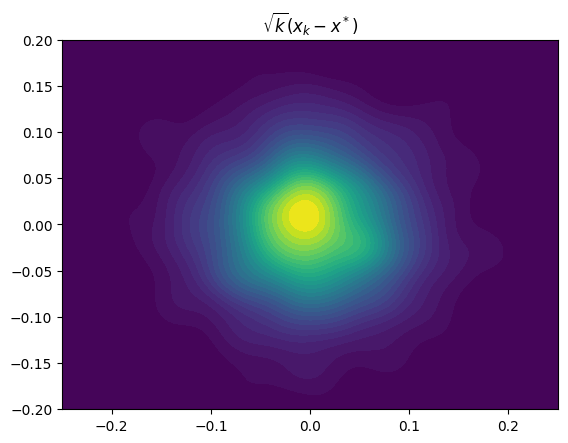}
    \caption{Setting 1}
    \label{fig:x_k}
\end{minipage}%
\begin{minipage}{.33\textwidth}
    \centering
    \includegraphics[width=1\textwidth]{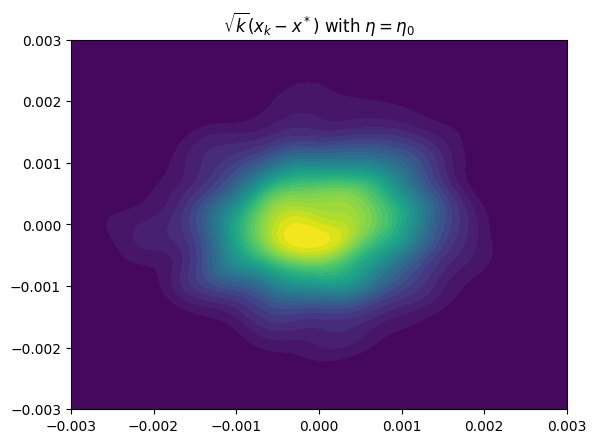}
    \caption{Setting 2}
    \label{fig:x_k_opt}
\end{minipage}%
\begin{minipage}{.33\textwidth}
    \centering
    \includegraphics[width=1\textwidth]{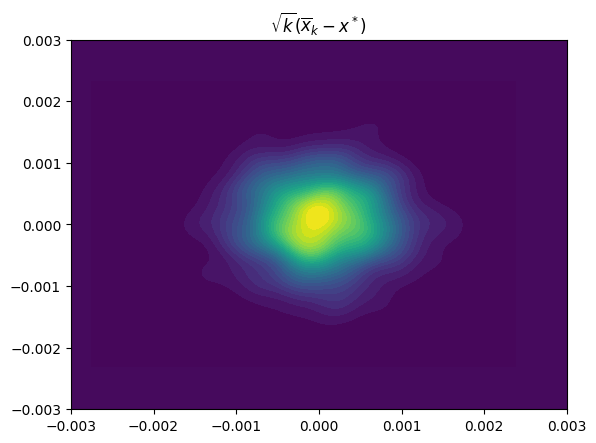}
    \caption{Setting 3}
    \label{fig:x_mean}
\end{minipage}%
\caption{Histograms of theorem-related values}
\end{figure}

The histograms for the three test settings are respectively presented in the figures \ref{fig:x_k}, \ref{fig:x_k_opt}, and \ref{fig:x_mean}.

\vspace{8pt}

It can be observed that selecting the optimal \(\eta = \eta_0\) results in a distribution that is significantly more concentrated around the single point $x^*$ ( take note of the differing scales across the figures). Additionally, the difference in covariance between \(\sqrt{k}(x_k - x^*)\) and \(\sqrt{k}(\overline{x}_k - x^*)\) is illustrated, providing empirical support for our theoretical results.

\section{Conclusion}
In this paper, we propose a new algorithm relying on the recently introduced concept of Stochastic Oracle Order~\cite{lobanov2024order} and the Polyak-Ruppert averaging technique for the stochastic minimization of strongly convex functions $f$. We then prove the asymptotic convergence of our method, by demonstrating in particular that the quantity $\sqrt{k}(\bar{x}-x_k) \sim \mathcal{N}(0,V)$, with a new and more accurate estimation of the matrix $V$ compared to the findings in~\cite{lobanov2024order}. 
Additionally, we derive optimal parameters that minimize the distribution around the minimizer \( x^\star \) of \( f \). We also present numerical experiments that support our theoretical results.
Future research directions include:
\begin{itemize}
\item Extending our method to achieve non-asymptotic convergence results,
\item Conducting further tests across a broader class of strongly convex functions and in higher dimensions.
\end{itemize}

\newpage
\bibliographystyle{unsrt}  
\bibliography{references}
\newpage
\appendix
\section{Appendix}\label{proofs}

\subsection{Proof of Lemma \ref{l1}}
Lemma \ref{l1} allows to apply the results from the work of B.T. Polyak and A.B. Juditsky~\cite{polyak1992acceleration} to the algorithm \ref{algorithm}. Proof of this Lemma consist of 4 parts, containing proofs of relevant Statements. 
\subsubsection{Proof of Statement \ref{st1}}
\proof
As $\mathbf{e}$ is a vector uniformly distributed on the Euclidean sphere, 

$\|\phi(x, y, \xi) \mathbf{e}\| = |\mathsf{sign}[f(x, \xi) - f(y, \xi)]| \cdot \| \mathbf{e}\| = 1 \leq K_1(1 + \|\xi\|)$.
\qed

\subsubsection{Proof of Statement \ref{st2}}
\proof
Let's prove all parts of this Statement step by step. 
\begin{enumerate}
    \item $\psi(x) = \mathbb{E}_{\xi} \frac{c}{\sqrt{d}}\frac{x + \xi}{\|x + \xi\|} = \int \frac{c}{\sqrt{d}} \frac{x + z}{\|x + z\|}dP(z)$ is defined.
    
    \item Let's check that $\psi^{\prime}(x) = \int \nabla_x \varphi(x + z) dP(z)$ for all $x \in X$, where $X = \{x: \|x\| < \varepsilon\}$,  $\varepsilon > 0, z \in \Omega$. 
    \begin{enumerate}
        \item\label{2.a} $\varphi(x + z) = \frac{c}{\sqrt{d}}\frac{x + z}{\|x + z\|}$ is a Lebesgue-integrable function of $z$ for each $x \in X$.
        \item For almost all $z \in \Omega$ the partial derivative $\nabla_x \varphi(x+z)$ exists for all $x \in X$ as
        
        \begin{equation*}
        \frac{\partial \varphi(x_i + z_i)}{\partial x_j} = 
         \begin{cases}
           -\frac{(x_i + z_i)(x_j + z_j)}{\|x+z\|^3} &\text{for $i \neq j$}\\
           \frac{1}{\|x+z\|} - \frac{(x_i+z_i)^2}{\|x+z\|^3} &\text{for $i = j.$}
         \end{cases}
        \end{equation*}

        \item\label{2.c} For all $i, j = \overline{1,d}$ there is an integrable function $\theta: \Omega \rightarrow \mathbb{R}$ such that $|\frac{\partial \varphi(x_i + z_i)}{\partial x_j}| \leq \theta(z)$ for all $x \in X$ and almost every $z \in \Omega$. 

        To demonstrate this, let's estimate the numerator of the partial derivative from above
        
        $$|x_i + z_i| \leq |x_i| + |z_i| \leq \|x\| + \|z\| < \varepsilon + \|z\|.$$

        and the denominator from below
        
        $$\|x+z\|^3 \geq (\|z\| - \|x\|)^3 = \|z\|^3 - 3\|z\|^2\|x\| + 3\|z\|\|x\|^2 - \|x\|^3 >$$
        
        $$ > \|z\|^3 - 3\varepsilon\|z\|^2 + 3\|z\|\|x\|^2 - \varepsilon^3 \geq \|z\|^3 - 3\varepsilon\|z\|^2 - \varepsilon^3.$$
        
        $$\|x + z\| \geq \|z\| - \|x\| > \|z\| - \varepsilon.$$
        
        Combining these inequalities, we obtain
        
        $$\left|-\frac{(x_i + z_i)(x_j + z_j)}{\|x+z\|^3}\right| < \frac{(\varepsilon + \|z\|)^2}{\|z\|^3 - 3\varepsilon\|z\|^2 - \varepsilon^3}.$$
        
        $$\left|\frac{1}{\|x+z\|} - \frac{(x_i+z_i)^2}{\|x+z\|^3}\right| \leq \frac{1}{\|x+z\|} + \left|\frac{(x_i+z_i)^2}{\|x+z\|^3}\right| < \frac{1}{\|z\| - \varepsilon} + \frac{(\varepsilon + \|z\|)^2}{\|z\|^3 - 3\varepsilon\|z\|^2 - \varepsilon^3}.$$
    \end{enumerate}
    Based on the above result, we can find the $\psi^{\prime}(0):$

    $\psi^{\prime}(0) = \int \nabla \varphi(z) dP(z) = \mathbb{E}_{\xi}\nabla \varphi(\xi) = \mathbb{E}_{\xi}[\frac{c}{\sqrt{d}}[\frac{1}{\|\xi\|}I - \frac{\xi \xi^T}{\|\xi\|^3}]] = \frac{c}{\sqrt{d}} [\mathbb{E}_{\xi} \frac{1}{\|\xi\|}I - \mathbb{E}_{\xi} \frac{\xi \xi^T}{\|\xi\|^3}] = $ 
    
    $ = \frac{c}{\sqrt{d}} [\mathbb{E}_{\xi} \frac{1}{\|\xi\|}I - \frac{1}{d} \mathbb{E}_{\xi} \frac{\|\xi\|^2}{\|\xi\|^3} I] = \frac{c}{\sqrt{d}} \int \|z\|^{-1} dP(z) (1 - \frac{1}{d}) I = \frac{c}{\sqrt{d}} \alpha (1 - \frac{1}{d}) I$, 
    
    where $\alpha = \int \|z\|^{-1} dP(z) < \infty$. 
    
    Here we use the facts that for $i \neq j \int \frac{z_i z_j}{\|z\|^3} dP(z) = 0$ due to spherical symmetry of $\xi$ and that $\int \frac{z_i^2}{\|z\|^3} dP(z) =\frac{1}{d} \int \frac{z_1^2 + \dots z_d^2}{\|z\|^3} dP(z) = \frac{1}{d} \int \frac{\|z\|^2}{\|z\|^3} dP(z).$

    \item From Assumption \ref{as2} we get that 

    $\psi(0) = \mathbb{E}_{\xi} \frac{c}{\sqrt{d}} \frac{\xi}{\|\xi\|} = 0$. 

    \item Let's estimate $\psi(x)=\mathbb{E}_\xi\frac{x + \xi}{\|x + \xi\|}$:

    Vector $x$ is a constant here, thus we will consider two components of a random vector $\xi = \xi_{\parallel}+\xi_{\perp}$: $\xi_{\parallel}\parallel x$ and $\xi_{\perp} \perp x$. Having that and from properties of mathematical expectation we can split the mathematical expectation we are looking for:
    
    $$
    \mathbb{E}_\xi\frac{x + \xi}{\|x+\xi\|}=\mathbb{E}_\xi\frac{x+\xi_\parallel+\xi_\perp}{\|x+\xi\|}=\mathbb{E}_\xi\frac{x+\xi_\parallel}{\|x+\xi\|}+\mathbb{E}_\xi\frac{\xi_\perp}{\|x+\xi\|}.
    $$

    Now we will find both terms of the sum.
    
    $$
    \mathbb{E}_\xi\frac{\xi_\perp}{\|x + \xi\|} = \int_{\mathbb{R}^d}\frac{z_{\perp}}{\|x+z_\parallel + z_\perp\|}dP(z_\parallel + z_\perp).
    $$
    
    From the fact that $\xi$'s distribution is spherically symmetrical, it follows that $dP(z)=dP(z_\parallel + z_\perp)=dP(z_\parallel - z_\perp)=dP(-z_\parallel + z_\perp)=dP(-z_\parallel - z_\perp)$.

    Firstly, due to this result we can calculate the integral only over half-space $\{z\in\mathbb{R}^d: x^Tz>0\}$:

    $$
    \int_{\mathbb{R}^d}\frac{z_{\perp}}{\|x+z_\parallel + z_\perp\|}dP(z_\parallel + z_\perp)=
    $$
    
    $$
    =\int_{\{x^Tz>0\}}\left[\frac{z_{\perp}}{\|x+z_\parallel + z_\perp\|}dP(z_\parallel + z_\perp)+\frac{-z_{\perp}}{\|x - z_\parallel - z_\perp\|}dP(-z_\parallel - z_\perp)\right]=
    $$

    \begin{equation}\label{intform1}
        =\int_{\{x^Tz>0\}}\left[\frac{z_{\perp}}{\|x+z_\parallel + z_\perp\|}+\frac{-z_{\perp}}{\|x - z_\parallel - z_\perp\|}\right]dP(z).
    \end{equation}

    This integral can also be calculated in other form:

    $$
    \int_{\mathbb{R}^d}\frac{z_{\perp}}{\|x+z_\parallel + z_\perp\|}dP(z_\parallel + z_\perp)=
    $$
    
    $$
    =\int_{\{x^Tz>0\}}\left[\frac{z_{\perp}}{\|x - z_\parallel + z_\perp\|}dP(-z_\parallel + z_\perp)+\frac{-z_{\perp}}{\|x + z_\parallel - z_\perp\|}dP(z_\parallel - z_\perp)\right]=
    $$

    \begin{equation}\label{intform2}
    =\int_{\{x^Tz>0\}}\left[\frac{z_{\perp}}{\|x-z_\parallel + z_\perp\|}+\frac{-z_{\perp}}{\|x + z_\parallel - z_\perp\|}\right]dP(z).
    \end{equation}

    Summing up the forms \ref{intform1} and \ref{intform2} and dividing the result by 2 we get:

    $$
    \int_{\mathbb{R}^d}\frac{z_{\perp}}{\|x+z_\parallel + z_\perp\|}dP(z_\parallel + z_\perp)=
    $$
    
    $$
    =\frac{1}{2}\left(\int_{\{x^Tz>0\}}\left[\frac{z_{\perp}}{\|x+z_\parallel + z_\perp\|}+\frac{-z_{\perp}}{\|x - z_\parallel - z_\perp\|}\right]dP(z) + \right.
    $$
    $$
    \left. +\int_{\{x^Tz>0\}}\left[\frac{z_{\perp}}{\|x-z_\parallel + z_\perp\|}+\frac{-z_{\perp}}{\|x + z_\parallel - z_\perp\|}\right]dP(z)\right)=
    $$
    $$
    =\frac{1}{2}\int_{\{x^Tz>0\}}\left[\frac{z_{\perp}}{\|x+z_\parallel + z_\perp\|}+\frac{-z_{\perp}}{\|x+z_\parallel - z_\perp\|}+\right.
    $$
    \begin{equation}\label{intfinal1}
    \left.=\frac{z_{\perp}}{\|x-z_\parallel + z_\perp\|}+\frac{-z_{\perp}}{\|x-z_\parallel - z_\perp\|}\right]dP(z).
    \end{equation}

    We use the 2-norm, thus $\|x + z_\parallel + z_\perp\| =\|x + z_\parallel - z_\perp\|$ and $\|x - z_\parallel + z_\perp\| =\|x - z_\parallel - z_\perp\|$.

    That means that the sum under the integral \ref{intfinal1} equals $0$.

    Hence, we obtain $\mathbb{E}_\xi\frac{x + \xi}{\|x + \xi\|} = \mathbb{E}_\xi\frac{x+\xi_\parallel}{\|x+\xi\|}$.
    Therefore:
    
    $$
    \mathbb{E}_\xi\frac{x+\xi_\parallel}{\|x+\xi\|}=\mathbb{E}_{\beta}\frac{x + \beta x}{\|x + \beta x\|},\ \beta\in\mathbb{R}.
    $$

    $\beta$'s distribution is determined by the distribution of $\xi$ as
    
    $$
    \rho_\beta\left(y\right) = \int_{\{x^Tz=0\}}\rho_\xi(\frac{yx}{\|x\|} + z)dP_\xi(z).
    $$

    $\xi$ is distributed spherically symmetrically, which means $\beta$ is also distributed symmetrically in $\mathbb{R}$.
    
    $$
    x^T\phi(x)=x^T\mathbb{E}_{\beta}\frac{x + \beta x}{\|x + \beta x\|}=x^T\int_{\mathbb{R}}\frac{x+yx}{\|x + yx\|}dP_\beta(y) = \frac{x^Tx}{\|x\|}\int_{\mathbb{R}}\frac{1+y}{|1 + y|}dP_\beta(y) = $$
    $$= \|x\|\left[-\int_{y<-1}dP_{\beta}(y) + \int_{y>-1}dP_{\beta}(y)\right] = \|x\|\cdot \underbrace{\left[P_\beta(\beta > -1) - P_\beta(\beta < -1)\right].}_{m}
    $$

    Recalling the fact of symmetrically distribution of $\beta$ and that it is induced by spherically symmetrical distribution (what implies that $P_\beta(-1 < \beta < 0) > 0$) we can assert that $P_\beta(\beta > -1) > P_\beta(\beta < -1))$, thus $m>0$. So:
    
    $$
    \forall x: \|x\| > 0\Rightarrow x^T\psi(x) = x^T\mathbb{E}_\xi\frac{x + \xi}{\|x+\xi\|} = m\frac{x^Tx}{\|x\|} = m\|x\| > 0.
    $$

    \item Let's check that $\psi^{\prime \prime}(x) = \int \nabla_x^2 \varphi(x + z) dP(z)$ for all $x \in X$, where $X = \{x: \|x\| < \varepsilon\}$,  $\varepsilon > 0, z \in \Omega$. 
    \begin{enumerate}
        \item $\nabla_x \varphi(x + z) = \frac{c}{\sqrt{d}}[\frac{1}{\|x + z\|}I - \frac{(x+z) (x+z)^T}{\|x+z\|^3}]$ is a Lebesgue-integrable function of $z$ for each $x \in X$ (proved in \ref{2.a}).
        \item For almost all $z \in \Omega$ the partial derivative $\nabla_x^2 \varphi(x+z)$ exists for all $x \in X$ as
        
        \begin{equation*}
        \frac{\partial}{\partial x_k}\left(-\frac{(x_i + z_i)(x_j + z_j)}{\|x+z\|^3}\right) = 
         \begin{cases}
           \frac{3(x_i+z_i)(x_j+z_j)(x_k+z_k)}{\|x+z\|^5} &\text{for $k \neq i$ and $k \neq j,$}\\
           -\frac{x_j+z_j}{\|x+z\|^3} + \frac{3(x_k+z_k)^2(x_j+z_j)}{\|x+z\|^5} &\text{for $k = i$ or $k = j.$}\\
           
         \end{cases}
        \end{equation*}
        
        \begin{equation*}
        \frac{\partial}{\partial x_k}\left(\frac{1}{\|x+z\|} - \frac{(x_i+z_i)^2}{\|x+z\|^3}\right) = 
         \begin{cases}
           -\frac{3(x_k+z_k)}{\|x+z\|^3} + \frac{3(x_k+z_k)^3}{\|x+z\|^5} &\text{for $k = i,$}\\
           -\frac{x_k+z_k}{\|x+z\|^3} + \frac{3(x_k+z_k)(x_i+z_i)^2}{\|x+z\|^5} &\text{for $k \neq i.$}\\
           
         \end{cases}
        \end{equation*}

        \item For all $i, j, k = \overline{1,d}$ there is an integrable function $\theta: \Omega \rightarrow \mathbb{R}$ such that 
        
        $|(\nabla_x^2 \varphi(x+z))_{ijk}| \leq \theta(z)$ for all $x \in X$ and almost every $z \in \Omega$. 

        Let's prove it. The necessary estimations for expressions $(x_i+z_i)$ and $\|x+z\|^3$ were obtained in \ref{2.c}. Thus, we should only estimate $\|x+z\|^5$:
        
        $$\|x+z\|^5 \geq (\|z\| - \|x\|)^5 = $$
        
        $$= \|z\|^5 - 5\|z\|^4\|x\| + 10\|z\|^3\|x\|^2 - 10\|z\|^2\|x\|^3 + 5\|z\|\|x\|^4 - \|x\|^5 > $$
        
        $$> \|z\|^5 - 5\varepsilon\|z\|^4 - 10\varepsilon^3\|z\|^2 - \varepsilon^5.$$
        
        Applying these inequalities as in \ref{2.c}, we gain the desired result. 
        
        So, we showed that $\psi^{\prime \prime}(x)$ exists in the neighbourhood of zero. Hence, by Teylor's theorem we have:
        $$\psi(x) = \psi(0) + \psi^{\prime}(0)x + O(\|x\|^2).$$
        
        Thus we obtain $$\|\psi^{\prime}(0)x - \psi(x)\| \leq K_2\|x\|^2 \leq K_2\|x\|^{1 + \lambda}$$ for $0 < \lambda \leq 1$.
        
    \end{enumerate}
    
\end{enumerate}
\qed

\subsubsection{Proof of Statement \ref{st3}}

\proof
$\chi(x) = \int \frac{c^2}{d} \frac{(x + z)(x+z)^T}{\|x + z\|^2}dP(z)$ is defined and is continuous at zero. 

$\chi(0) = \frac{c^2}{d} \mathbb{E}_{\xi} \frac{\xi \xi^T}{\|\xi\|^2} = \frac{c^2}{d} \cdot \frac{1}{d} \mathbb{E}_{\xi} \frac{\|\xi\|^2}{\|\xi\|^2} \cdot I = \frac{c^2}{d^2}I.$
\qed

\subsubsection{Proof of Statement \ref{st4}}
\proof
$-G = -\frac{c}{\sqrt{d}} \alpha (1 - \frac{1}{d}) \nabla^2 f(x^*)$ is Hurwitz because $\nabla^2 f(x^*) > 0$ (Assumption \ref{as1}).  
\qed

Thus, all Statements are fulfilled. 
\subsection{Proof of Theorem \ref{t1}}

We give a proof of Theorem \ref{t1}, which is based on the work of B.T. Polyak and A.B. Juditsky~\cite{polyak1992acceleration}. This theorem provides a new asymptotic estimate for the convergence rate of the algorithm \ref{algorithm}. Our reasoning is similar to the proof of Theorem 3 in the above-mentioned work.

\proof Let us check whether the assumptions of Theorem 2~\cite{polyak1992acceleration} are fulfilled. For that purpose, we transform the algorithm \ref{algorithm} in the following way:

\begin{equation}\label{trans_alg}
\begin{gathered}
x_{k+1} = x_k - \eta_k R(x_k) + \eta_k R(x_k) - \eta_k \phi(x_k + \gamma \mathbf{e}, x_k - \gamma \mathbf{e}, \xi_k)\mathbf{e} = \\
= x_k - \eta_k R(x_k) + \eta_k(R(x_k) - \phi(x_k + \gamma \mathbf{e}, x_k - \gamma \mathbf{e}, \xi_k)\mathbf{e}) = \\
= x_k - \eta_k R(x_k) + \eta_k \xi_k(x_k - x^*);
\end{gathered}
\end{equation}

here 

\begin{equation}
\begin{gathered}
R(x_k) = \mathbb{E}_{\xi_1, \mathbf{e}} [\phi(x_k + \gamma \mathbf{e}, x_k - \gamma \mathbf{e}, \xi_k)\mathbf{e}] = \mathbb{E}_{\xi_1, \mathbf{e}} [\mathsf{sign} [\langle \nabla f(x_k, \xi_k), \mathbf{e} \rangle]\mathbf{e}] = \\
= \mathbb{E}_{\xi_1} \frac{c}{\sqrt{d}} \frac{\nabla f(x_k, \xi_k)}{\|\nabla f(x_k, \xi_k)\|} = 
\mathbb{E}_{\xi_1} \frac{c}{\sqrt{d}} \frac{\nabla f(x_k) + \xi_k}{\|\nabla f(x_k) + \xi_k\|} = \\
= \int \varphi(\nabla f(x_k) + z) dP(z) =\psi(\nabla f(x_k))
\end{gathered}
\end{equation}

\begin{equation}\label{xi}
\xi_k(x_k - x^*) = R(x_k) - \phi(x_k + \gamma \mathbf{e}, x_k - \gamma \mathbf{e}, \xi_k)\mathbf{e}
\end{equation}

From Statement \ref{st2} of Lemma \ref{l1} we have that $R^T(x)\nabla f(x) > 0$ for all $x \neq 0$. Let $f(x^*) = 0$ for the sake of simplicity. It follows from Assumptions \ref{as1} and Statement \ref{st2} of Lemma \ref{l1} that there exist $\alpha > 0, \alpha^{\prime} > 0, \varepsilon > 0$ such that 
$$R^T(x)\nabla f(x) \geq \alpha \|\nabla f(x)\|^2 \geq \alpha^{\prime} f(x)$$

for all $\|x - x^*\| \leq \varepsilon$; hence $f(x)$ is a Lyapunov function for \ref{trans_alg}, and all corresponding conditions of Theorem 2~\cite{polyak1992acceleration} are fulfilled.

So we obtain by Statement \ref{st2} of Lemma \ref{l1} that

$$\|R(x) - G(x - x^*)\| = \|\psi(\nabla f(x)) - \psi^{\prime}(0) \nabla^2 f(x^*)(x - x^*)\| \leq $$
$$ \leq \|\psi(\nabla f(x)) - \psi^{\prime}(0) \nabla f(x)\| + \|\psi^{\prime}(0) \nabla f(x) - \psi^{\prime}(0) \nabla^2 f(x^*)(x - x^*)\| \leq $$
$$ \leq K\|\nabla f(x)\|^{1+\lambda} + \|\psi^{\prime}(0)\| \cdot \|\nabla f(x) - \nabla^2 f(x^*) (x - x^*)\| \leq$$
$$\leq K\|x - x^*\|^{1 + \lambda} + K\|x - x^*\|^{2} \leq K\|x - x^*\|^{1 + \lambda}.$$

Hence Assumption 3.2 of Theorem 2~\cite{polyak1992acceleration} is fulfilled.

Next, using the notation $\Delta_k$ for the error of equation \ref{algorithm} ($\Delta_{k} = x_k - x^*$), we note that $\xi_k(\Delta_k)$ is a martingale-difference process and that
$$\mathbb{E} \|\xi_k(\Delta_k)\|^2 \leq K(1 + \|\Delta_k\|^2).$$

So, as concluded in the proof of the Theorem 2~\cite{polyak1992acceleration} (see Parts 1 and 2), $\Delta_k \rightarrow 0$ and for every $\varepsilon > 0$, there exists some $R < \infty$ such that $P(\sup_{k}\|\Delta_k\| \leq R) \geq 1 - \varepsilon$. Define the stopping time $\tau_R = \inf\{k \geq 1: \|\Delta_k\| > R\}$. From Part 2 Theorem 2~\cite{polyak1992acceleration} we have

\begin{equation}\label{tau}
    \mathbb{E} \|\Delta_k\|^2 I(k \leq \tau_R) \leq K \eta_k.
\end{equation}

Then, from \ref{tau} and Statements \ref{st2} and \ref{st3} of Lemma \ref{l1}, we have that 
$$\|\mathbb{E}(\xi_k(\Delta_k) \xi_k(\Delta_k)^T | \mathcal{F}_k) - \chi(0)\| \leq K\|\chi(\Delta_k) - \chi(0)\| + K\|\Delta_k\|^2 \rightarrow 0.$$

Next, we obtain that 
$$\mathbb{E}(\|\xi_k(\Delta_k)\|^2 I(\|\xi_k(\Delta_k)\| > C)| \mathcal{F}_k) \leq K\mathbb{E}(\|\xi_k\|^2 I(\|\xi_k(\Delta_k)\| > C)| \mathcal{F}_k) + K\|\Delta_k\|^2.$$

From the definition \ref{xi} by Statement \ref{st1} of Lemma \ref{l1}, we get that 
$$I(\|\xi_k(\Delta_k)\| > C) \leq I(\|\Delta_k\| > KC) + I(\|\xi_k\| > KC);$$

so with $k \rightarrow \infty$

$$ \mathbb{E} (\|\xi_k(\Delta_k)\|^2 I(\|\xi_k(\Delta_k)\| > C) | \mathcal{F}_k) \leq o(1) + K \mathbb{E} (\|\xi_k\|^2 I(\|\xi_k\| > KC) | \mathcal{F}_k) \rightarrow 0.$$

(Here $o(1) \rightarrow 0$ as $t \rightarrow \infty$). This means that Assumption 3.3 of Theorem 2~\cite{polyak1992acceleration} holds. Therefore all conditions of the proposition of Theorem 2~\cite{polyak1992acceleration} are fulfilled, and the matrix $V$ is defined by the equation

\begin{equation}
\begin{gathered}
    V = G^{-1} \chi(0) (G^{-1})^T = \\
    = \left(\frac{c}{\sqrt{d}} \alpha (1 - \frac{1}{d}) \nabla^2 f(x^*)\right)^{-1} \frac{c^2}{d^2}I \left(\frac{c}{\sqrt{d}} \alpha (1 - \frac{1}{d}) \nabla^2 f(x^*)\right)^{-1} = \\
    = \frac{d}{(d-1)^2\alpha^2}\nabla^2f(x^*)^{-2}.
\end{gathered}
\end{equation}
\qed
\subsection{Calculation of optimal $\eta$ for the matrix \ref{eq_gasnikov} (Proof of Lemma \ref{lem:opt_eta})}\label{sec_opt_eta}














In this section we find the optimal $\eta$ for asymptotic covariance matrix \ref{eq_gasnikov}. Knowledge about optimal $\eta$ is necessary in order to be able to compare above-mentioned matrix with asymptotic covariance matrix \ref{eq_ours}.

\proof
Consider the minimization of an unitarily invariant norm of $$V=\frac{\eta^2}{d}\left( 2\eta(1-\frac{1}{d})\frac{c}{\sqrt{d}}\alpha\nabla^2f\left( 
x^* \right) - I \right)^{-1},$$ namely the spectral norm. Denote $A := 2(1-\frac{1}{d})\frac{c}{\sqrt{d}}\alpha\nabla^2f\left( 
x^* \right)$. In this way,

$$\|V\| = \frac{\eta^2}{d}\|(\eta A - I)^{-1}\|.$$

Let's understand what $\|(\eta A - I)^{-1}\|$ is equal to. For that purpose we leverage the singular value decomposition of the matrix $\eta A - I$. $$\eta A - I = U\Sigma V^T,$$ 

where $U, V$ are unitary matrices, $\Sigma$ is diagonal matrix with singular values of the matrix $\eta A - I$ in its diagonal. Thus, 

$$\|(\eta A - I)^{-1}\| = \|(U\Sigma V^T)^{-1}\| = \|V \Sigma^{-1} U^{T}\| = \|\Sigma^{-1}\| = \frac{1}{\sigma_d(\eta A - I)},$$

where $\sigma_d$ is the minimum singular value. 

Since the matrix $\eta A - I$ is real, symmetric and positive definite we obtain that 

$$\sigma_d(\eta A - I) = \sqrt{\lambda_d((\eta A - I)^* (\eta A - I))} = \sqrt{\lambda_d((\eta A - I)^2)} =$$ $$=\sqrt{(\lambda_d(\eta A - I))^2} = \lambda_d(\eta A - I).$$

The matrix $A$ is normal, so it's unitary diagonalizable. It means that we can represent the matrix $A$ as $Q\Lambda Q^{-1}$, where $Q$ is unitary matrix, $\Lambda$ is diagonal matrix. Hence,

$$\lambda_d(\eta A - I) = \lambda_d(Q (\eta \Lambda - I) Q^{-1}) = \eta \lambda_d - 1,$$

where $\lambda_d$ is a minimum eigenvalue of the matrix $A$.

So, we obtained that $g(\eta) := \|V\| = \frac{\eta^2}{d}\cdot \frac{1}{\eta \lambda_d - 1}.$ Let's minimize it by $\eta$.

$$g(\eta)^{\prime} = \frac{2\eta}{d}\cdot\frac{1}{\eta \lambda_d - 1} - \frac{\eta^2}{d} \cdot \frac{\lambda_d}{(\eta \lambda_d - 1)^2} = \frac{2\eta(\eta \lambda_d - 1) - \eta^2 \lambda_d}{d (\eta \lambda_d - 1)^2}.$$

$$2\eta(\eta \lambda_d - 1) - \eta^2 \lambda_d = 0.$$

$$\eta(\eta\lambda_d - 2) = 0.$$

We are interested in non-trivial solution, so the optimal $\eta$ is
$$\eta_0 = \frac{2}{\lambda_d}.$$

Now, let's find the $\lambda_d$:

$$\lambda_d = \lambda_d(A) = \lambda_d\left(2(1-\frac{1}{d})\frac{c}{\sqrt{d}}\alpha\nabla^2f\left( 
x^* \right)\right) = 2(1-\frac{1}{d})\frac{c}{\sqrt{d}}\alpha \lambda_d\left(\nabla^2f(x^*)\right).$$

Since the function $f$ is $\mu$-strongly convex, $\lambda_i\left(\nabla^2f(x^*)\right) \geq \mu > 0$ $\forall i = \overline{1, d},$ so we can choose such $\mu$ that $\lambda_d\left(\nabla^2f(x^*)\right) = \mu$. 

Thus, we obtained the optimal $\eta$:

\begin{equation}\label{eq_opt_eta}
    \eta_0 = \frac{d\sqrt{d}}{(d-1)c\alpha \mu}.
\end{equation}

\qed

\subsection{Comparison of covariance matrices (Proof of Lemma \ref{lem:better-estimation})}

In this section, we compare our obtained results with those presented in the work of Lobanov et al.~\cite{lobanov2024order}. We demonstrate that our asymptotic covariance matrix $V_{our}$ (\ref{eq_ours}) is not greater (in the matrix sense) than the asymptotic covariance matrix $V_{other}$ (\ref{eq_gasnikov}) with the optimal $\eta$ (\ref{eq_opt_eta}). To establish this, we prove the positive semi-definiteness of the difference $V_{other} - V_{our}$.

\proof
Recall the formulas, obtained earlier:

$$V_{other} = \frac{\eta^2}{d}\left( 2\eta(1-\frac{1}{d})\frac{c}{\sqrt{d}}\alpha\nabla^2f\left( 
x^* \right) - I \right)^{-1},$$
$$V_{our} = \frac{d}{(d-1)^2\alpha^2}\nabla^2f(x^*)^{-2}.$$

First, substitute the optimal $\eta$ (\ref{eq_opt_eta}) into the formula for $V_{other}$ and get:

$$V_{other} = \frac{d^2}{(d-1)^2c^2\alpha^2\mu^2}\left(\frac{2}{\mu}\nabla^2f(x^*) - I\right)^{-1}.$$

Denote $A := \left(\frac{2}{\mu}\nabla^2f(x^*) - I\right)^{-1}$, $B := \nabla^2f(x^*)^{-2}.$

To demonstrate the positive semi-definiteness of a matrix, it is sufficient to show that its eigenvalues are non-negative. For this purpose, we present the difference $V_{other} - V_{our}$ in a more convenient way. 

Let's start with $\nabla^2f(x^*)$. Since this matrix is normal, it's unitary diagonalizable. It means that we can represent the matrix $\nabla^2f(x^*)$ as $U\Lambda U^{-1}$, where $U$ is unitary matrix, $\Lambda$ is diagonal matrix. In this way,

$$A = U (\frac{2}{\mu} \Lambda - I)^{-1}U^{-1},$$
$$B = U \Lambda^{-2} U^{-1}.$$

Write down the difference of covariance matrices.

$$V_{other} - V_{our} = U \left( \frac{d^2}{(d-1)^2c^2\alpha^2\mu^2} (\frac{2}{\mu} \Lambda - I)^{-1} - \frac{d}{(d-1)^2\alpha^2} \Lambda^{-2} \right) U^{-1}.$$

Find the eigenvalues $m_i$, $i=\overline{1,d}$ for the matrix $V_{other} - V_{our}.$

$$m_i = \frac{d^2}{(d-1)^2c^2\alpha^2\mu^2}\cdot \frac{1}{\frac{2}{\mu}\lambda_i - 1} - \frac{d}{(d-1)^2\alpha^2} \cdot \frac{1}{\lambda_i^2},$$

where $\lambda_i$ is the $i$'th eigenvalue of the matrix $\nabla^2f(x^*)$. 

Simplify the expression, we obtain:

$$m_i = \frac{d\mu(d\lambda_i^2-2c^2\lambda_i\mu+c^2\mu^2)}{(d-1)^2\alpha^2c^2\mu^2(2\lambda_i - \mu)\lambda_i^2}.$$

Since $d \geq 1$ as dimension of space; $\mu > 0$ as a constant of strong convexity; $c \leq 1$ (Lemma \ref{l52}); $\lambda_i \geq \mu$ as $f$ is $\mu$-strongly convex function; we have:

$$m_i \geq \frac{d\mu(\lambda_i^2-2c^2\lambda_i\mu+c^2\mu^2)}{(d-1)^2\alpha^2c^2\mu^2(2\lambda_i - \mu)\lambda_i^2} = \frac{d\mu(c^2(\mu-\lambda_i)^2 + \lambda_i^2(1-c^2))}{(d-1)^2\alpha^2c^2\mu^2(2\lambda_i - \mu)\lambda_i^2} \geq 0.$$

Therefore, we got that all eigenvalues of $V_{other} - V_{our}$ are non-negative, so the matrix is positive semi-definite. 

\qed
\label{article_end}

\end{document}